\documentclass[journal,compsoc]{IEEEtran}
\makeatletter
\let\NAT@parse\undefined

\makeatother

\usepackage{graphicx}
\usepackage{amsmath,amssymb,amsfonts,bbding,amsthm,pifont}
\usepackage{color,xcolor}         
\usepackage{textcomp}
\usepackage{multirow,colortbl,booktabs,threeparttable}
\usepackage{enumitem,subfigure}

\usepackage{algorithm}
\usepackage[noend]{algpseudocode}
\usepackage[square,numbers,sort&compress]{natbib}
\definecolor{revision}{HTML}{2F70AF} 
\usepackage[colorlinks,linkcolor=black,anchorcolor=black,citecolor=black]{hyperref}

\newtheorem{lemma}{Lemma}[section]

\usepackage{dcolumn}

\newcommand{\T}{\mathrm{T}}
\newcommand{\D}{\mathrm{D}}
\newcommand{\RH}{\mathrm{H}}

\definecolor{abl}{HTML}{74709b}

\def\BibTeX{{\rm B\kern-.05em{\sc i\kern-.025em b}\kern-.08em
    T\kern-.1667em\lower.7ex\hbox{E}\kern-.125emX}}
\begin{document}
\author{Anonymous Authors }
\title{DeepFilter: A Transformer-style Framework for Accurate and Efficient Process Monitoring}

\author{Hao Wang, Zhichao Chen$^\dagger$, Licheng Pan, Xiaoyu Jiang, Yichen Song, Qunshan
 He, Xinggao Liu$^\dagger$
\thanks{This work is supported by the National Natural Science Foundation of China (62073288, 12075212), China Postdoctoral Science Foundation (2025M781449) and the National Key Research and Development Program of China (Grant No. 2021YFC2101100). } 
\thanks{Hao Wang, Zhichao Chen, Licheng Pan, Xiaoyu Jiang, Yichen Song, Qunshan
 He, and Xinggao Liu are with the State Key Laboratory of Industrial Control Technology, College of Control Science and Engineering, Zhejiang University, Hangzhou 310027, China (e-mail: haohaow@zju.edu.cn; 
 12032042@zju.edu.cn;
 22132045@zju.edu.cn;
 jiangxiaoyu@zju.edu.cn
 syc\_1203@zju.edu.cn;
 heqs@zju.edu.cn;
 lxg@zju.edu.cn).
}}


\maketitle

\begin{abstract}

Effective process monitoring is increasingly vital in industrial automation for ensuring operational safety, necessitating both high accuracy and efficiency. Although Transformers have demonstrated success in various fields, their canonical form based on the self-attention mechanism is inadequate for process monitoring due to two primary limitations: \ding{182} the step-wise correlations captured by the self-attention mechanism struggle to capture discriminative patterns in monitoring logs due to the lack of semantics of each step, thus compromising accuracy; \ding{183} the quadratic computational complexity of self-attention hampers efficiency. To address these issues, we propose DeepFilter, which modifies the Transformer for process monitoring. The core modification is replacing the self-attention layer with an efficient filtering layer that excels at capturing long-term discriminative patterns with reduced complexity. Equipped with the efficient filtering layer, DeepFilter enhances both accuracy and efficiency, meeting the stringent demands of process monitoring. Experimental results on real-world process monitoring datasets validate DeepFilter’s superiority in terms of accuracy and efficiency compared to existing state-of-the-art models.
\end{abstract}

\def\abstractname{Impact statement}
\begin{abstract}

The process monitoring task is characterized by stringent demands for accuracy and efficiency. Current transformer-based methods, characterized by self-attention for temporal fusion, exhibit limitations in accurately understanding the semantic context and efficiently processing monitoring logs, rendering them inadequate for process monitoring. To address these limitations, we introduce DeepFilter, which revises the self-attention mechanism to improve both accuracy and efficiency. As a straightforward yet versatile approach, DeepFilter provides an instrumental baseline for practitioners in process monitoring, whether initiating new projects or enhancing existing capabilities.
\textit{Interested readers can implement DeepFilter following the code in the supplementary file.} The monitoring logs are available and daily updated on \url{https://data.rmtc.org.cn/gis/PubIndex.html}.

\end{abstract}



\section{Introduction}
\label{sec:introduction}

\IEEEPARstart{M}{onitoring} the quality variable through advanced time-series analysis techniques is paramount for ensuring operational safety in industrial automation~\cite{10854556,10770754}. These techniques are widely employed in process engineering~\cite{9691915}, manufacturing~\cite{chen2024knowledgenn,10411940}, and energy conversion~\cite{wang2024taiattentionmixer}, where the growing demands for higher efficiency and cost-effectiveness often drive equipment to operate under extreme conditions, thereby increasing the likelihood of catastrophic failures~\cite{wang2024taiattentionmixer,xu2024denoising}. For example, in chemical engineering, the Haber-Bosch process for ammonia synthesis requires temperatures above 400°C and pressures exceeding 200 bar, escalating risks of equipment failure and hazardous leaks~\cite{10663500}. Similarly, nuclear power plants operate reactors at high pressures and temperatures to maximize output, heightening the threat of catastrophic radiation leakage~\cite{safe1}. These examples underscore the critical role of process monitoring systems for safeguarding operations in industrial automation~\cite{yang2022paradigm,yao2020industrial}.

Process monitoring operates by firstly estimating next-step values of the quality variable from historical monitoring logs, and subsequently identifying anomalies by detecting significant deviations between predicted and observed values~\cite{wang2024taiattentionmixer,yang2022paradigm}.
An instrumental process monitoring system must satisfy the dual requirements of high \textbf{accuracy} and computational \textbf{efficiency} for safeguarding industrial equipments.
For example, in large-scale chemical plants~\cite{10663500,chen2024ocndplvm}, precise estimation of temperature and pressure is necessary to detect their subtle shifts, enabling operators to intervene before potential reactor instabilities escalate. Meanwhile, real-time efficiency ensures these corrective measures are executed promptly, minimizing the risk of safety incidents or costly downtime.

To achieve accurate and efficient process monitoring, a variety of data-driven algorithms have been developed, evolving from early identification methods to advanced deep learning models. Traditional methods such as ARIMA~\cite{Chandrakar2017} offer computational simplicity but are limited in handling the nonlinear complexity of industrial data. Statistical methods, including decision trees~\cite{rf}, XGBoost~\cite{xgb}, and regression-based approaches~\cite{Chan2017}, provide improved accuracy through richer feature extraction but rely heavily on manual engineering and struggle to scale with large datasets. In contrast, deep learning architectures—spanning convolutional~\cite{wang2020remaining}, recurrent~\cite{zhang2018,Choi2020}, and graph neural networks~\cite{9612030,9740531}—enable automated feature extraction and GPU acceleration, driving both performance and speed. Notably, Transformers~\cite{Vaswani2017} have emerged as an exemplar solution, characterized by the self-attention layer for capturing temporal patterns. This layer dynamically computes weighted dependencies across all time steps in the monitoring logs, capturing step-wise relationships useful for prediction~\cite{yuan2023attention}. Additionally, the architecture is optimized for GPU acceleration and is well-suited for large-scale monitoring data~\cite{Vaswani2017} processing. These advancements have made Transformers a preferred choice in industrial time-series analytics~\cite{yuan2023attention,yuan2024quality,qiu2025duet}\footnote{The standard Transformer consists of encoder and decoder components. In this paper, "Transformer" refers to the encoder component commonly used in time-series analytics, rather than the full encoder-decoder architecture.}.

Despite the promise of Transformers, we contend that the self-attention mechanism exhibits two fundamental limitations that restrict its suitability for the stringent demands of process monitoring. \ding{182}~\textbf{The reliance on step-wise correlations results in suboptimal accuracy for process monitoring}, since it hinders the modeling of long-term discriminative patterns. Specifically, discriminative patterns in monitoring logs—such as gradual drifts, recurring trends, or periodic fluctuations—often emerge at the sub-sequence level and may require modeling long-term histories. It would be difficult to model them by calculating step-wise relationships in a learned space. Moreover, individual sensor observations in industrial monitoring logs carry limited semantic meaning, further hindering the model's ability to extract discriminative patterns necessary for accurate monitoring. \ding{183}~\textbf{The reliance on step-wise correlations results in suboptimal efficiency for process monitoring.}
Each self-attention layer requires pairwise interactions across all time steps, leading to quadratic computational and memory costs. 
Collectively, canonical Transformers fall short of delivering the required accuracy and efficiency in process monitoring.

To overcome these limitations, we propose \textit{DeepFilter}, a refined Transformer that replaces the self-attention layer with a novel efficient filtering layer specifically tailored for process monitoring. This layer performs adaptive filtering across the entire temporal sequence, effectively modeling long-term discriminative patterns. Theoretical analysis further confirms its capacity to enhance representations of long-term discriminative patterns. Moreover, by eliminating the quadratic complexity inherent in self-attention, the efficient filtering layer significantly reduces computational overhead. Extensive evaluations on real-world datasets demonstrate that \textit{DeepFilter} consistently delivers superior accuracy and efficiency relative to state-of-the-art models, highlighting its role as an instrumental approach for Transformer-based process monitoring.

\textbf{Organization.}
 Section~\ref{sec:Background} reviews related works and highlights technical gaps.
 Section~\ref{sec:pre} introduces the preliminary concepts essential to our methodology.
 Section~\ref{sec:proposed} provides a detailed description of the DeepFilter architecture. Section~\ref{sec:experiment} presents a case study on process monitoring in real-world nuclear power plants, showcasing DeepFilter's accuracy and efficiency improvements. Finally, we summarize our conclusions and limitations, and outline directions for future research.

\section{Related Works}\label{sec:Background}

In the era of Industry 4.0, reliability has emerged as a critical aspect in algorithm design and equipment maintenance~\cite{wang2024tifsescm,wang2025toiswbm,li2024nipsremoving}. Alongside tasks like soft sensing~\cite{pantmoe,pancmoe,chen2025blending}, fault detection~\cite{wu2024catch,11026875}, and data recovery~\cite{wang2024taselsptd,wang2025tnnlspot,chen2024newimp}, process monitoring serves as a vital safeguarding strategy by detecting early signs of faults or outliers~\cite{10854556,10770754}. Modern process monitoring approaches are broadly classified into three categories: identification methods, statistical methods, and deep methods.

Early identification methods, such as Auto-Regressive (AR), Moving Average (MA), and their integrated forms (ARIMA)\cite{Chandrakar2017}, offer computational simplicity and real-time processing. However, their effectiveness is limited by an inability to capture nonlinear temporal dependencies. In contrast, statistical methods such as decision trees~\cite{xgb} and generalized linear models~\cite{Chan2017} can model nonlinear patterns, but they typically require intensive manual feature engineering and face scalability challenges in large-scale industrial applications.
The advent of deep learning has revolutionized process monitoring by enabling automatic feature extraction and enhanced parallel computing. Various architectures, such as Convolutional Neural Networks (CNNs) \cite{wang2020remaining}, Recurrent Neural Networks (RNNs) \cite{zhang2018,Choi2020}, and Graph Neural Networks (GNNs) \cite{9740531,9612030}, have been developed to extract discriminative representations from monitoring logs for next-value prediction. For example, a spatiotemporal attention-based RNN model \cite{yuan2020deep} has been proposed to capture the nonlinearity among process variables and their temporal dynamics; a multi-scale attention-enhanced CNN~\cite{yuan2024quality} has been proposed to identify long- and short-term patterns; a multi-scale residual CNN~\cite{liu2022model} has been proposed to extract high-dimensional nonlinear features at multiple scales. Building on these advances, Transformers~\cite{Vaswani2017} have gained prominence for their scalability and parallel processing capabilities~\cite{yuan2023attention,yuan2024quality}. Initial applications employed self-attention to model step-wise relationships within monitoring logs~\cite{zhang2021bmt,pu2022mvstt,yuan2024quality}. Subsequent works enhanced attention mechanisms~\cite{zhou2021informer,liu2021pyraformer} to improve modeling accuracy and efficiency. Recognizing that step-wise correlations often lack semantic richness in industrial logs, recent studies have shifted towards modeling variate-wise correlations~\cite{wang2024taiattentionmixer,liuitransformer} and statistics for prediction~\cite{yue2025nipsOLinear}.

In the broad field of time-series modeling~\cite{wu2025k2vae,AutoCTS++}, spectral methods have become a cornerstone for improving both model accuracy and computational efficiency of neural network architectures~\cite{yi2025survey,wang2025iclrpswi,wu2021autoformer}. By decomposing temporal signals into their constituent frequency components~\cite{piao2024fredformer,ye2024frequency}, neural networks can more effectively capture salient dynamics~\cite{li2025ftmixer}, including seasonal (CoST~\cite{woocost}), low-frequency (FiLM~\cite{zhou2022film}), and periodic (TimesNet~\cite{timesnet}) patterns. Such frequency-enhanced representations have proven particularly beneficial for enhancing Transformer-based architectures~\cite{wu2021autoformer,zhou2022fedformer}. Beyond augmenting neural network architectures, Fourier-based techniques have also been leveraged for objective debiasing~\cite{wang2025iclrfredf,li2025nipstowards}, data compression~\cite{xu2020learning}, and data augmentation~\cite{zhang2022self}, further highlighting the versatility of spectral analysis as a fundamental paradigm in modern time-series modeling.

\section{Preliminaries}\label{sec:pre}
\subsection{Problem Definition}

Monitoring logs consist of chronological sequences of observations $\left[ S(1), S(2), \ldots, S(\mathrm{P}) \right]$, where each $S(t) \in \mathbb{R}^\mathrm{1 \times D_{\mathrm{in}}}$ represents the observation at the $t$-th step with $D_{\mathrm{in}}$ covariates. We define T as the historical window length, H as the prediction horizon. At an arbitrary time step $t$, the historical sequence is represented as $\mathbf{X} = \left[ S(t - \mathrm{T} + 1), \ldots, S(t) \right]$, and the corresponding quality variable $y^{\mathrm{H}} \in \mathbb{R}$ is specified as the last feature in $S(t+\mathrm{H})$.

The objective of process monitoring is to develop a predictive model $g: \mathbb{R}^\mathrm{T \times D_{\mathrm{in}}} \rightarrow \mathbb{R}$ that generates the quality variable prediction $\hat{y}^{\mathrm{H}} = g(\mathbf{X})$. Since the training dataset predominantly comprises normal operational logs, anomalies can be detected as significant deviations between the actual and predicted quality variable values.

\subsection{Fourier Transform}
The Discrete Fourier Transform (DFT) is a fundamental tool in digital signal processing, enabling the conversion of discrete time-domain data into its spectral representation~\cite{wang2025iclrfredf,oppenheim1999discrete}. This transformation is pivotal for analyzing the periodic components inherent in a signal.

Let $\mathbf{x} = [x_0, x_1, \ldots, x_{\mathrm{T}-1}]$ represent a real-valued time series of length $\mathrm{T}$.
The DFT maps this sequence to a sequence of complex numbers $\mathbf{x}^\mathrm{(F)}=[\mathbf{x}^\mathrm{(F)}_0,...,\mathbf{x}^\mathrm{(F)}_\mathrm{T-1}]$. Each component $\mathbf{x}^\mathrm{(F)}_\omega$ corresponds to the frequency $2\pi \omega/\mathrm{T}$ and is defined as:
\begin{equation*}
    \mathbf{x}^\mathrm{(F)}_\omega=\sum_{t=0}^\mathrm{\mathrm{T}-1} \mathbf{x}_te^{-j2\pi \omega t/\mathrm{T}}=[\mathbf{x}_0,..,\mathbf{x}_\mathrm{\mathrm{T}-1}]\cdot\left[\begin{array}{c}
\mathbf{F}_{0,\omega}  \\
\mathbf{F}_{1,\omega}  \\
  \\
\mathbf{F}_{\mathrm{T-1},\omega}  \\
\end{array}\right],
\end{equation*}
    where $j$ is the imaginary unit ($j^2=-1$), $\mathbf{F}_{t,\omega}:=e^{-j2\pi \omega t/\mathrm{T}}$. This transform can be concisely expressed in matrix form as:
    \begin{equation*}
    \begin{aligned}
\mathbf{x}^\mathrm{(F)}&=\mathbf{x}\cdot \mathbf{F}\\&=[\mathbf{x}_0,..,\mathbf{x}_\mathrm{n-1}]\cdot\left[\begin{array}{ccc}
\mathbf{F}_{0,0}  &\cdots& \mathbf{F}_{0,\mathrm{T}-1}  \\
\mathbf{F}_{1,0}  &\cdots& \mathbf{F}_{1,\mathrm{T}-1} \\
 \vdots &\ddots&\vdots \\
\mathbf{F}_{\mathrm{T-1},0}   &\cdots& \mathbf{F}_{\mathrm{T-1},\mathrm{T}-1}\\
\end{array}\right].
    \end{aligned}
    \end{equation*}

The DFT is an invertible transformation. The original time-domain sequence $\mathbf{x}$ can be perfectly reconstructed from its spectral representation $\mathbf{X}$ using the Inverse DFT (IDFT):
\begin{equation}
    \mathbf{x}_t=\sum_{\omega=0}^\mathrm{\mathrm{T}-1} \mathbf{x}^\mathrm{(F)}_\omega e^{j2\pi \omega t/\mathrm{T}}, t=0,1,...,\mathrm{T-1}.
\end{equation}

The direct computation of the DFT has a time complexity of $\mathcal{O}(\mathrm{T}^2)$, which is inefficient for long sequences. However, the Fast Fourier Transform (FFT) algorithm significantly reduces this complexity to $\mathcal{O}(\mathrm{T} \log \mathrm{T})$ by exploiting the symmetric properties of the DFT matrix, making large-scale frequency analysis efficient and scalable in real-world applications~\cite{oppenheim1999discrete}.

\section{Methodology}\label{sec:proposed}

\subsection{Motivation}

While Transformers are widely used for time-series analysis~\cite{zhou2021informer,wen2023transformers}, they exhibit fundamental limitations for real-world process monitoring, which demands both high accuracy and efficiency. Specifically, the self-attention mechanism exhibits limitations in modeling the long-term temporal dependencies characteristic of industrial processes, limiting accuracy, while the architecture's quadratic complexity is prohibitive for real-time deployment, hampering efficiency\footnote{Although techniques like model sparsification~\cite{zhang2022monitoring,wang2023high,sparseformer} have been proposed to improve efficiency, they often compromise accuracy, failing to resolve the inherent trade-off between performance and computational cost.}.

To address these challenges, spectral methods present a promising alternative, with successful applications in machine vision~\cite{rao2021global,xu2020learning}, audio processing~\cite{lancucki2021fastpitch}, and general time-series modeling~\cite{xu2023fits,wang2025iclrfredf,fbm}. However, their application in process monitoring remains nascent, typically confined to using FFT as a simple preprocessing step for feature extraction~\cite{liao2021manufacturing} rather than integrating adaptive filtering into an end-to-end deep learning architecture.

Our study is therefore motivated by two observations in process monitoring: the limitations of standard Transformers~\cite{wang2024taiattentionmixer} and the underexplored potential of spectral methods. In this section, we introduce DeepFilter, an architecture that modernizes the standard Transformer for process monitoring. The core innovation is replacing the Transformer's self-attention module with an efficient filtering layer for capturing discriminative temporal patterns accurately and efficiently. This design enhances monitoring accuracy and efficiency, making DeepFilter an instrumental approach in process monitoring tasks.

\subsection{Global Filtering Block}

\begin{figure}
    \centering
    \includegraphics[width=\linewidth]{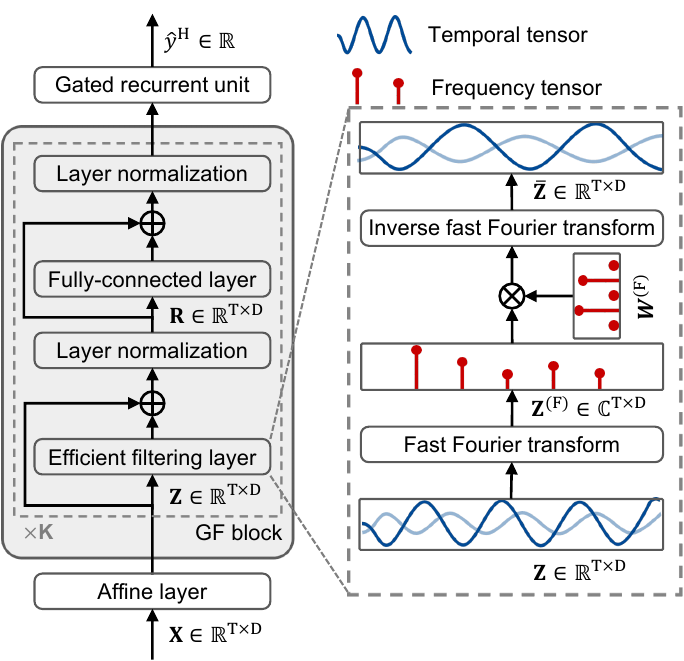}
    \caption{Overview of the core components in DeepFilter.}
    \label{fig:model}
\end{figure}

The fundamental component of DeepFilter is the Global Filtering (GF) block. As depicted in \autoref{fig:model}, it retains the overall structure of Transformer blocks but substitutes the self-attention layer with a unique efficient filtering layer.

Suppose $\mathbf{Z} \in \mathbb{R}^{\mathrm{T} \times \mathrm{D}}$ is the input to an arbitrary $k$-th GF block, where $\mathrm{T}$ is the length of the historical window and $\mathrm{D}$ is the hidden dimension. The efficient filtering layer first transforms $\mathbf{Z}$ from the time domain to the frequency domain:
\begin{equation}
    \mathbf{Z}^{(\mathrm{F})} = \mathcal{F}(\mathbf{Z}),
\end{equation}
where $\mathcal{F}$ denotes the FFT operation, applied independently along the time axis for each hidden dimension. Notably, we employ the real FFT instead of the canonical FFT. The real FFT preserves only the right-sided spectrum symmetric to the left-sided one, ensuring that the inverse transform yields real-valued results, which is essential for compatibility with established PyTorch modules. The obtained $\mathbf{Z}^{(\mathrm{F})}$ has several advantageous properties. For example, it enables the isolation and removal of high-frequency noise \cite{highnoise1} while highlighting potentially discriminative low-frequency features \cite{highnoise2}. 

To extract discriminative patterns and suppress noise from $\mathbf{Z}^{(\mathrm{F})}$, inspired by \cite{wang2025iclrfredf,rao2021global}, we apply a filtering operation using the Hadamard product:
\begin{equation}\label{eq:barzkf}
    \bar{\mathbf{Z}}^{(\mathrm{F})} = \mathbf{Z}^{(\mathrm{F})} \odot \mathbf{W}^{(\mathrm{F})},
\end{equation}
where $\mathbf{W}^{(\mathrm{F})} \in \mathbb{C}^{\mathrm{T}' \times \mathrm{D}}$ consists of learnable parameters optimized during training to identify discriminative patterns. The filtered sequence is subsequently transformed back to the time domain via the inverse (real) FFT:
\begin{equation}\label{eq:barzk}
    \bar{\mathbf{Z}} = \mathcal{F}^{-1}(\bar{\mathbf{Z}}^{(\mathrm{F})}),
\end{equation}
followed by a residual connection and layer normalization to stabilize training and mitigate gradient degradation:
\begin{equation}
    \mathbf{R} = \mathrm{LayerNorm}(\bar{\mathbf{Z}} + \mathbf{Z}),
\end{equation}

Although the efficient filtering layer adaptively captures the dominant temporal patterns in each channel, it does not incorporate channel-wise interactions. To fill this gap, we introduce a fully connected layer as follows:
\begin{equation}\label{eq:r}
    \mathrm{FC}(\mathbf{R}) = \mathrm{ReLU}(\mathbf{R} \mathbf{W}^{(1)} + \mathbf{b}^{(1)}) \mathbf{W}^{(2)} + \mathbf{b}^{(2)},
\end{equation}
\begin{equation}\label{eq:barr}
    \bar{\mathbf{R}} = \mathrm{LayerNorm}(\mathrm{FC}(\mathbf{R}) + \mathbf{R}),
\end{equation}
where $\mathbf{W}^{(1)}$, $\mathbf{b}^{(1)}$, $\mathbf{W}^{(2)}$, and $\mathbf{b}^{(2)}$ are learnable parameters. To stabilize the training process, residual connection and layer normalization are subsequently applied.
In summary, the GF block captures temporal patterns and channel-wise interactions via the efficient filtering layer and the fully connected layer, respectively. The output $\bar{\mathbf{R}}\in \mathbb{R}^{\mathrm{T} \times \mathrm{D}}$ is a representation that comprehensively encodes the historical sequence.

\subsection{Theoretical Analysis}\label{sec:theory}
In this section, we perform theoretical analysis on the efficient filtering layer. The analysis demonstrates that the efficient filtering layer excels at both accuracy and efficiency, meeting the dual stringent demands of process monitoring. 

First, we discuss the improvement in monitoring accuracy, which is attributed to two abilities of the efficient filtering layer.
\ding{182}~\textbf{Capturing long-term discriminative patterns.} According to Lemma~\ref{thm:conv}, the layer is equivalent to a circular convolution between the historical sequence and a large convolution kernel, with the kernel size equal to the historical window length $\mathrm{T}$. This large kernel size allows for the modeling of long-term sequence-wise dependencies, which are discriminative for process monitoring, as opposed to the step-wise correlations captured by standard Transformers. \ding{183}~\textbf{Obtaining decorrelated representations.} According to Lemma~\ref{thm:ortho}, the representations obtained via FFT consistently become decorrelated. It enhances monitoring accuracy by minimizing redundancy and correlation among dimensions, thus eliminating the need for the model to capture these correlations, which in turn reduces model complexity and improves generalization.

\begin{lemma}\label{thm:conv}
    Suppose $\mathbf{W}=\mathcal{F}^{-1}(\mathbf{W}^\mathrm{(F)})$, "$*$" is the circular convolution operator, the filtered sequence in \eqref{eq:barzk} can be acquired via circular convolution below~\cite{oppenheim1999discrete}
    \begin{equation*}
        \bar{\mathbf{Z}} = \mathbf{W} * \mathbf{Z}.
    \end{equation*}
\end{lemma}
\begin{proof}
    It is equivalent to prove $\mathcal{F}(\mathbf{W} * \mathbf{Z}) = \mathbf{Z}^{(\mathrm{F})} \odot \mathbf{W}^{(\mathrm{F})}$. To this end, we express the $n$-th element of $\bar{\mathbf{Z}}$ as 
        $\bar{\mathbf{Z}}_n=\sum_{m=0}^\mathrm{T-1} \mathbf{W}_m \mathbf{Z}_{(n-m)\% \mathrm{T}}$.
    On this basis, the FFT of $\bar{\mathbf{Z}}$ is denoted as $\bar{\mathbf{Z}}^\mathrm{(F)}$, where the $\omega$-th element is given by:
    
    \begin{equation*}
    \begin{aligned}
    \bar{\mathbf{Z}}^\mathrm{(F)}_\omega 
    & =\sum_{n=0}^{\mathrm{T}-1} \sum_{m=0}^\mathrm{T-1} \mathbf{W}_m \mathbf{Z}_{(n-m)\% \mathrm{T}}  e^{-\frac{2\pi i}{T}n\omega } \\
    & =\sum_{n=0}^{\mathrm{T}-1}  \sum_{m=0}^\mathrm{T-1} \mathbf{W}_m e^{-\frac{2\pi i}{T}m\omega } \mathbf{Z}_{(n-m)\% \mathrm{T}}  e^{-\frac{2\pi i}{T}\omega (n-m)} \\
    &=\sum_{m=0}^\mathrm{T-1} \mathbf{W}_m e^{-\frac{2\pi i}{T}m\omega }  \sum_{n=0}^{\mathrm{T}-1}   \mathbf{Z}_{(n-m)\% \mathrm{T}}  e^{-\frac{2\pi i}{T}\omega (n-m)} \\
    &= \mathbf{W}^\mathrm{(F)}_\omega \sum_{n=0}^{\mathrm{T}-1}   \mathbf{Z}_{(n-m)\% \mathrm{T}}  e^{-\frac{2\pi i}{T}\omega (n-m)} \\
    &= \mathbf{W}^\mathrm{(F)}_\omega  \sum_{n=m}^{\mathrm{T}-1}   \mathbf{Z}_{n-m}  e^{-\frac{2\pi i}{T}\omega (n-m)}\\
    &+ \mathbf{W}^\mathrm{(F)}_\omega  \sum_{n=0}^{m-1}   \mathbf{Z}_{n-m+\mathrm{T}}  e^{-\frac{2\pi i}{T}\omega (n-m)}\\
    &= \mathbf{W}^\mathrm{(F)}_\omega  (\sum_{n=0}^{\mathrm{T}-m-1}   \mathbf{Z}_{n}  e^{-\frac{2\pi i}{T}\omega (n)}+ \sum_{n=\mathrm{T}-m}^{\mathrm{T}-1}   \mathbf{Z}_{n}  e^{-\frac{2\pi i}{T}\omega (n)})\\
    &= \mathbf{W}^\mathrm{(F)}_\omega  \cdot \mathbf{Z}^\mathrm{(F)}_\omega .
    \end{aligned}
    \end{equation*}
    
    Thus, the equation $\mathcal{F}(\mathbf{W} * \mathbf{Z}) = \mathbf{Z}^{(\mathrm{F})} \odot \mathbf{W}^{(\mathrm{F})}$ holds, and the proof is thereby completed.
\end{proof}

\begin{lemma}[Modified from \cite{wang2025iclrfredf}]\label{thm:ortho}
Let \(\mathbf{x}\) be a zero-mean, wide-sense stationary sequence of length \(T\).  
As \(T \to \infty\), its FFT coefficients \(\mathbf{x}^\mathrm{(F)}_\omega\) satisfy  
\[
\lim_{T \to \infty} \mathbb{E}\!\left[\mathbf{x}^\mathrm{(F)}_\omega \bigl(\mathbf{x}^\mathrm{(F)}_{\omega'}\bigr)^*\right] = S_x(\omega)\,\delta_{\omega\omega'},
\]  
where \(\mathcal{F}\{\mathbf{x}\} = \mathbf{x}^\mathrm{(F)}\), \(S_x(\omega)\) is the power spectral density, and \(\delta_{\omega\omega'}\) is the Kronecker delta.
\end{lemma}
\begin{table}
\caption{Complexity Analysis}
\label{tab:complex}
\centering\renewcommand{\arraystretch}{1}\small\setlength{\tabcolsep}{1.5mm}
\begin{tabular}{llll}
    \toprule
    Blocks               &Complexity                &Sequential Ops                    &Path Length\\
    \midrule
    RNN & $\mathcal{O}(\T\cdot\D^2)$ & $\mathcal{O}(\T)$ & $\mathcal{O}(\T)$ \\
    Transformer& $\mathcal{O}(\T^2\cdot\D)$ & $\mathcal{O}(1)$ & $\mathcal{O}(1)$\\
    iTransformer& $\mathcal{O}(\T\cdot\D^2)$ & $\mathcal{O}(1)$ & $\mathcal{O}(1)$\\
    DeepFilter& $\mathcal{O}(\T\cdot\log\T\cdot\D)$ & $\mathcal{O}(1)$ & $\mathcal{O}(1)$\\
    \bottomrule
  \end{tabular}
\end{table}

Second, we discuss the improvement in monitoring efficiency by analyzing and comparing the computational costs of the temporal fusion mechanisms in DeepFilter and baselines~\cite{Vaswani2017,liuitransformer}. Specifically, we evaluate the computational costs of transforming a sequence \((\mathbf{x}_1,...,\mathbf{x}_\T)\) into another sequence \((\mathbf{z}_1,...,\mathbf{z}_\T)\) of equal length \(\T\) and feature dimension \(\D\). We consider three key metrics~\cite{Vaswani2017} in \autoref{tab:complex}: complexity (measured by the number of floating-point operations), sequential operations (the number of non-parallelizable operations), and path length (the minimum number of layers required to model relationships between any two time steps).
\ding{182}~\textbf{Most existing deep models suffer from various sources of inefficiency.} RNNs exhibit inefficiency due to heavy sequential operations and lengthy path lengths. By reducing the sequential operations and path lengths, Transformer and iTransformer can process entire sequences within a single computational block and leverage GPU acceleration effectively. However, they incur quadratic complexity relative to T and D, respectively, resulting in inefficiency when dealing with long sequences or numerous covariates.
\ding{183}~\textbf{DeepFilter substantially enhances theoretical efficiency.} Specifically, the efficient filtering layer reduces the computational complexity to \(\mathcal{O}(\T \cdot \log \T \cdot \D)\), significantly lower than that of self-attention layers. Meanwhile, it retains the minimum sequential operations and path length, which makes it suitable for GPU acceleration and free from excessive cascading.

\subsection{DeepFilter Architecture and Learning Objective}
The GF block efficiently processes historical monitoring logs and excels at capturing discriminative temporal patterns. However, this block focuses on encapsulating these patterns into a comprehensive representation $\bar{\mathbf{R}}$, without generating the predicted value of the quality variable for process monitoring. To bridge this gap, we introduce DeepFilter, which integrates cascaded GF blocks to achieve process monitoring.

The architecture of DeepFilter is illustrated in \autoref{fig:model}. It begins by transforming the historical monitoring log $\mathbf{X} \in \mathbb{R}^{\mathrm{T} \times \mathrm{D}_{\mathrm{in}}}$ into a latent representation through an affine layer:
\begin{equation}
    \mathbf{Z}^{0} = \mathbf{X}\mathbf{W}^{(0)}+\mathbf{b}^{(0)}
\end{equation}
where $\mathbf{Z}^{0} \in \mathbb{R}^{\mathrm{T} \times \mathrm{D}}$ represents the initial embeddings with hidden dimension $\mathrm{D}$, and $\mathbf{W}^{(0)}$ and $\mathbf{b}^{(0)}$ are learnable parameters. These embeddings are then sequentially processed through $\mathrm{K}$ GF blocks. Let $\mathbf{Z}^{k}$ denote the input to the $k$-th GF block; the output of this block is given by:
\begin{equation}
    \bar{\mathbf{R}}^{k} := \operatorname{GF}_k(\mathbf{Z}^{k}),
\end{equation}
where $\operatorname{GF}_k(\cdot)$ performs the transformations from Eq \eqref{eq:barzkf} to \eqref{eq:barr} sequentially. The output of each GF block serves as the input for the subsequent block:
\begin{equation}
    \mathbf{Z}^{k+1} := \bar{\mathbf{R}}^{k}.
\end{equation}

The output of the final GF block, $\bar{\mathbf{R}}^{\mathrm{K}}$, which provides a comprehensive temporal representation of the monitoring logs, is processed by a Gated Recurrent Unit (GRU). The GRU aggregates information across the time steps, where its final hidden state, which summarizes the entire sequence, is passed through a linear projection layer to produce the prediction for the quality variable:
\begin{equation}
    \hat{y}^{\mathrm{H}} := \mathrm{GRU}(\bar{\mathbf{R}}^\mathrm{K}),
\end{equation}
where $\hat{y}^{\mathrm{H}}$ is the prediction for the quality variable with prediction horizon $\mathrm{H}$.

The learnable parameters in DeepFilter are optimized by minimizing the mean squared error (MSE) between the predicted and actual quality variable:
\begin{equation}
    \mathcal{L} := \left( y^{\mathrm{H}} - \hat{y}^{\mathrm{H}} \right)^2.
\end{equation}

\section{Experiments and Results}\label{sec:experiment}
In this section, we empirically validate the effectiveness of DeepFilter in the context of process monitoring. Specifically, four research problems are investigated.

\begin{enumerate}[leftmargin=*]
    \item \textbf{Accuracy:} \textit{Does DeepFilter achieve superior performance?} Section \ref{sec:overall} compares the accuracy of DeepFilter and baselines on real-world process monitoring datasets.
    \item \textbf{Efficacy:} \textit{How does it work?} Section \ref{sec:ab} performs an ablation study to discern the contribution of its components.
    \item \textbf{Efficiency:} \textit{Does it operate efficiently?} Section \ref{sec:complex} evaluates the running time under various settings.
    \item \textbf{Sensitivity:} \textit{Is it sensitive to hyperparameters?} Appendix \ref{sec:param} evaluates DeepFilter given different hyperparameter values. 
\end{enumerate}

\subsection{Dataset Collection and Background}
\begin{figure}
    \centering
    \subfigure[National station distribution.]{\includegraphics[width=0.48\columnwidth]{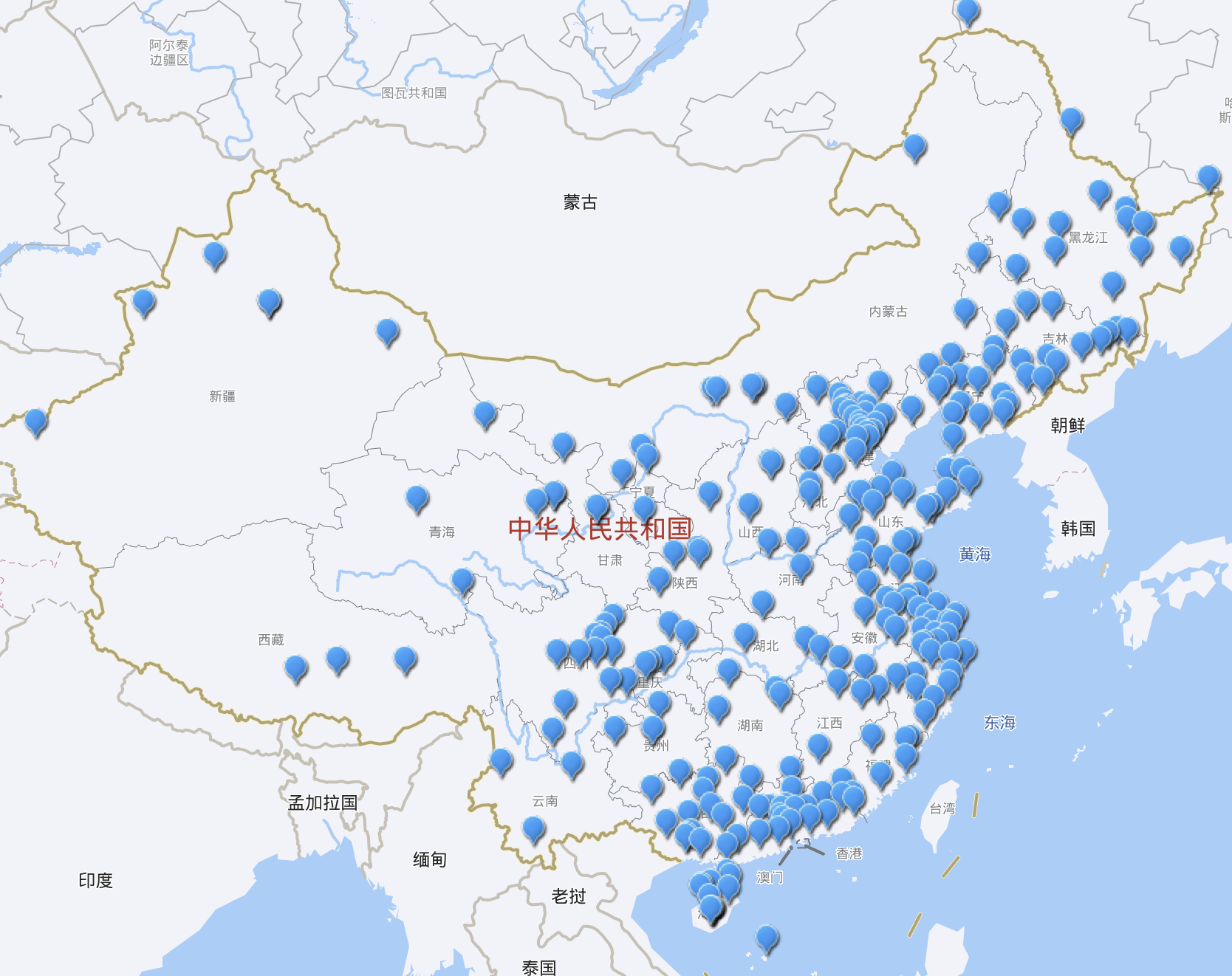}}\quad
    \subfigure[Composition of a single station.]{\includegraphics[width=0.465\columnwidth]{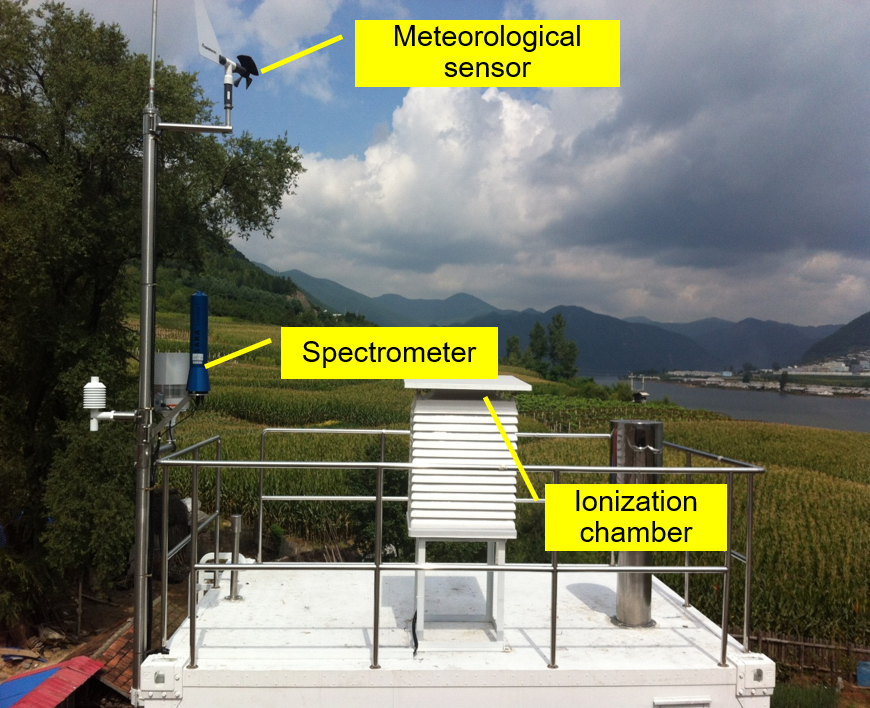}}
    \caption{Overview of the nuclear monitoring stations in our project.}
    \label{stationChart}
\end{figure}

\begin{table}
    \setlength{\tabcolsep}{6pt}\renewcommand{\arraystretch}{1}\small
    \caption{The quality and process variables in the monitoring log.}
    \label{dataTab}
    \centering
    
      \begin{tabular}{ll}
        \toprule
        \textbf{Field}               & \textbf{Unit of measure}\\
        \midrule
        \textbf{Atmospheric radiation}\\
        Dose rate           & nGy / h\\
        Spectrometer measures (1024 channels)       & ANSI/IEEE N42.42\\
        \midrule
        \textbf{Meteorological conditions}\\
        Temperature           & $^{\circ}$C\\
        Humidity       & \%\\
        Atmosphere pressure           & hPa\\
        Wind direction       & Clockwise angle\\
        Wind speed           & m / s\\
        Amount of precipitation      & mm\\
        \midrule
        \textbf{Spectrometer operating conditions}\\
        Battery voltage     & V\\
        Spectrometer voltage        & V\\
        Spectrometer temperature         & $^{\circ}$C\\
        \bottomrule
      \end{tabular}
    
\end{table}


\begin{figure*}
    \centering
    \subfigure[The dynamics of the quality variable: dose rate.]{\includegraphics[width=0.49\linewidth]{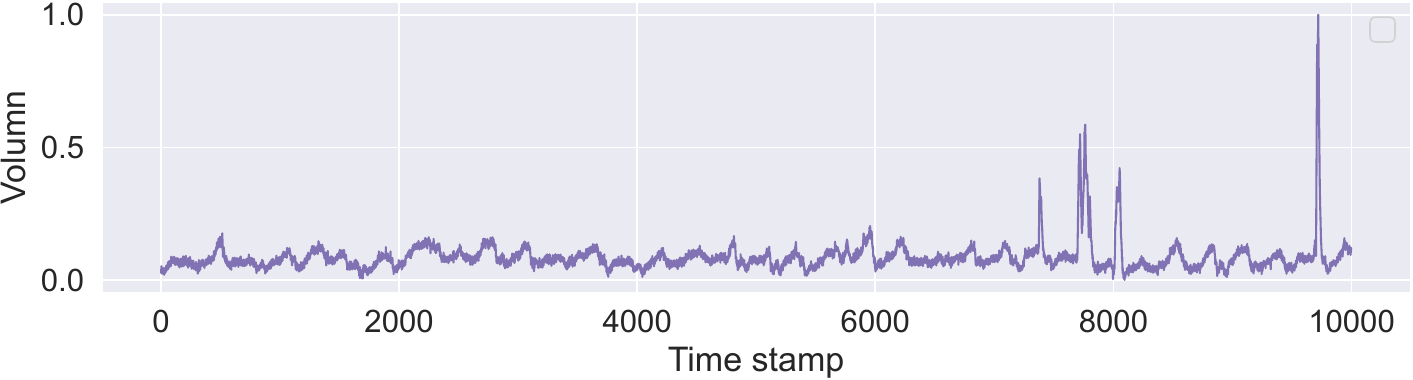}}
    \subfigure[The 64th channel of the spectrometer measurements and its FFT.]{\includegraphics[width=0.49\linewidth]{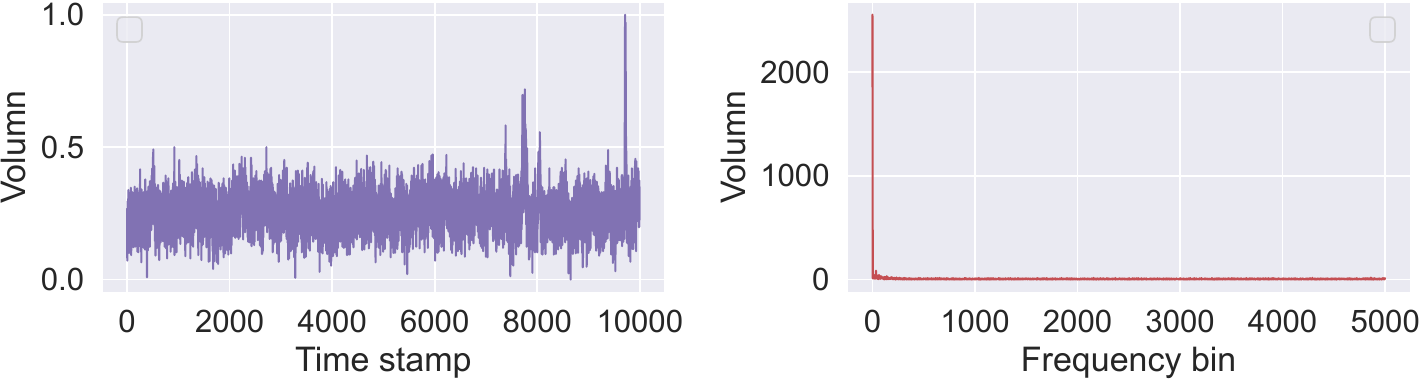}}
    \subfigure[The temperature measurements and its FFT.]{\includegraphics[width=0.49\linewidth]{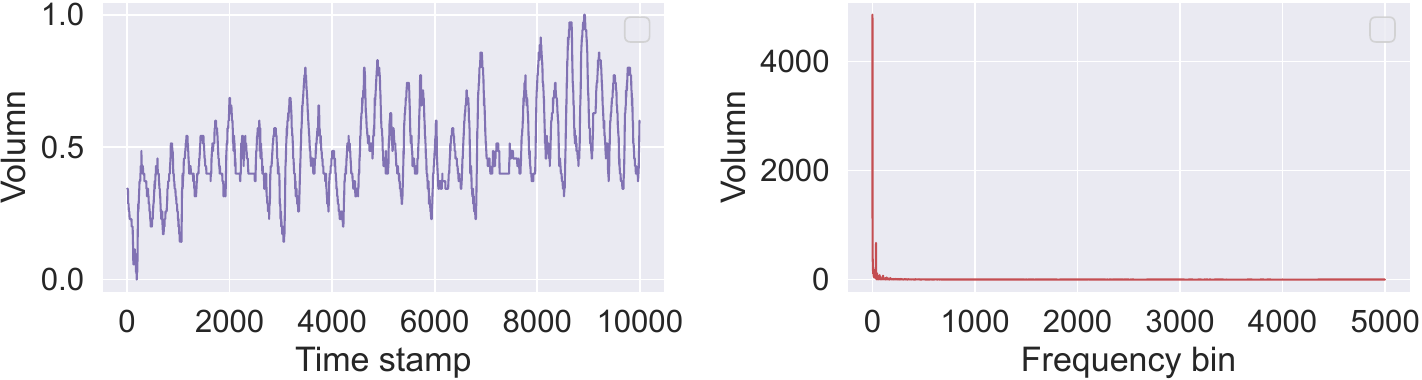}}
    \subfigure[The humidity measurements and its FFT.]{\includegraphics[width=0.49\linewidth]{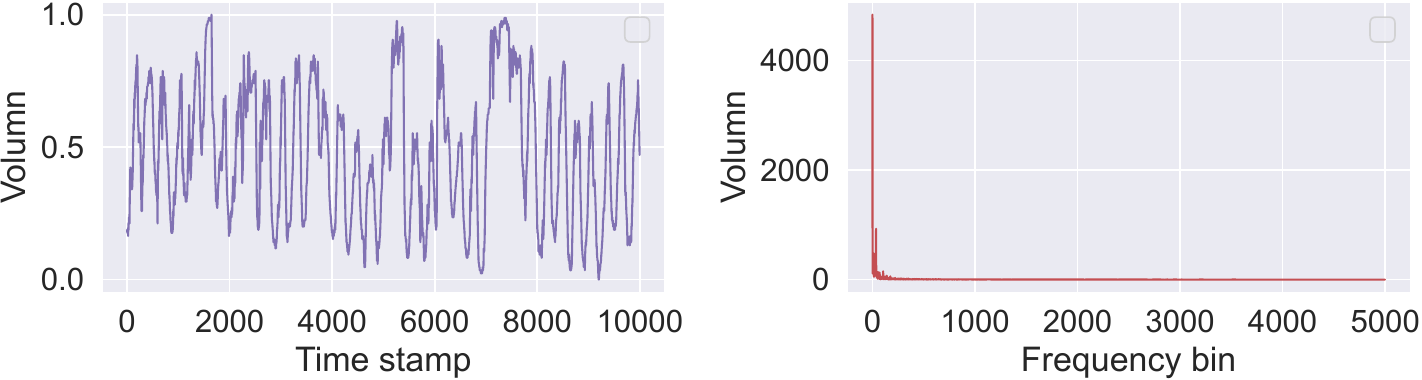}}
    \caption{Dynamics of the quality variable (a) and some important process variables (b-d) in the collected dataset from Jinan station. We also provide the amplitude characteristics of their FFT in (b-d).}
    \label{dataFig}
\end{figure*}

Nuclear power plants play a crucial role in industrial automation, supplying stable and efficient energy. However, they pose security risks, as operational anomalies can cause radionuclide leaks, leading to severe environmental pollution and casualties~\cite{safe1}. To address these risks, automated monitoring networks have been deployed globally, including the RadNet in the United States~\cite{li2020unconventional}, the Fixed Point Surveillance Network in Canada~\cite{liu2018development}, and the Atmospheric Nuclear Radiation Monitoring Network (ANRMN) in China, as shown in \autoref{stationChart}. These systems enable continuous, reliable monitoring of radionuclide concentrations, reflecting whether the nuclear power plants are operating normally.

The key quality variable monitored in ANRMN is the atmospheric $\gamma$-ray dose rate, measured by ionization chambers~\cite{weather1}. \autoref{dataFig}(a) illustrates the non-stationary dynamics of this variable, likely due to external factors such as weather conditions. To account for these influences, the monitoring systems also include additional process variables like spectrometer measurements (1024 channels), meteorological conditions (e.g., precipitation), and operational parameters (e.g., battery voltage). These variables display diverse temporal patterns, with frequency domain analysis revealing concentrated energy in low-frequency bands, diminishing at higher frequencies (see \autoref{dataFig}). This highlights the potential of frequency-based models to extract meaningful representations.

The final monitoring logs integrate 1,024-channel spectrometer data, 6 meteorological covariates, and 3 operational parameters as summarized in~\autoref{dataTab}. These monitoring logs provide a comprehensive view of operational status, enabling the construction of effective process monitoring systems for nuclear power plants.
    
\subsection{Experimental Setup}

\begin{table*}
    \caption{Comparative Study on the Hegang and Jinan datasets over four forecast horizons (\(\RH=1,3,5,7\)).}
    \label{comparativeTab}
    \begin{threeparttable}
    \setlength{\tabcolsep}{1.7mm}\centering
    \begin{tabular}{lcccccccccccccccc}
    \toprule
    {\multirow{2}{*}{Methods}} & \multicolumn{2}{c}{H=1} && \multicolumn{2}{c}{H=3} && \multicolumn{2}{c}{H=5} && \multicolumn{2}{c}{H=7}\\
    \cmidrule{2-3}\cmidrule{5-6}\cmidrule{8-9}\cmidrule{11-12}
    &  MAE & R$^2$ && MAE & R$^2$ && MAE & R$^2$ && MAE & R$^2$ &\\ 
    \hline
   \rowcolor{blue!8} \multicolumn{13}{l}{\textbf{Dataset: Hegang Station}} \\
AR & 0.135$^*_{\pm 0.000}$ & 0.064$^*_{\pm 0.000}$ & &0.137$^*_{\pm 0.000}$ & -0.038$^*_{\pm 0.000}$ & &0.140$^*_{\pm 0.000}$ & -0.111$^*_{\pm 0.000}$ & &0.141$^*_{\pm 0.000}$ & -0.163$^*_{\pm 0.000}$  \\
MA & 0.133$^*_{\pm 0.000}$ & 0.110$^*_{\pm 0.000}$ & &0.136$^*_{\pm 0.000}$ & -0.042$^*_{\pm 0.000}$ & &0.138$^*_{\pm 0.000}$ & -0.096$^*_{\pm 0.000}$ & &0.140$^*_{\pm 0.000}$ & -0.158$^*_{\pm 0.000}$  \\
ARIMA & 0.135$^*_{\pm 0.000}$ & 0.063$^*_{\pm 0.000}$ & &0.148$^*_{\pm 0.000}$ & -0.246$^*_{\pm 0.000}$ & &0.147$^*_{\pm 0.000}$ & -0.238$^*_{\pm 0.000}$ & &0.143$^*_{\pm 0.000}$ & -0.256$^*_{\pm 0.000}$  \\
LASSO & 0.045$^*_{\pm 0.000}$ & 0.282$^*_{\pm 0.000}$ & &0.046$^*_{\pm 0.000}$ & 0.243$^*_{\pm 0.000}$ & &0.047$^*_{\pm 0.000}$ & 0.183$^*_{\pm 0.000}$ & &0.049$^*_{\pm 0.000}$ & 0.109$^*_{\pm 0.000}$  \\
SVR & 0.023$^*_{\pm 0.000}$ & 0.890$^*_{\pm 0.000}$ & &0.034$^*_{\pm 0.000}$ & 0.779$^*_{\pm 0.000}$ & &0.051$^*_{\pm 0.000}$ & 0.540$^*_{\pm 0.000}$ & &0.058$^*_{\pm 0.000}$ & 0.391$^*_{\pm 0.000}$  \\
XGBoost & 0.016$^*_{\pm 0.000}$ & 0.943$^*_{\pm 0.000}$ & &0.018$^*_{\pm 0.000}$ & 0.902$^*_{\pm 0.000}$ & &0.022$^*_{\pm 0.000}$ & 0.838$_{\pm 0.000}$ & &0.023$_{\pm 0.000}$ & 0.778$_{\pm 0.000}$  \\
LSTM & 0.021$^*_{\pm 0.001}$ & 0.846$^*_{\pm 0.011}$ & &0.023$^*_{\pm 0.001}$ & 0.790$^*_{\pm 0.013}$ & &0.026$^*_{\pm 0.001}$ & 0.718$^*_{\pm 0.015}$ & &0.027$^*_{\pm 0.002}$ & 0.651$^*_{\pm 0.024}$  \\
GRU & 0.023$^*_{\pm 0.001}$ & 0.832$^*_{\pm 0.008}$ & &0.025$^*_{\pm 0.001}$ & 0.773$^*_{\pm 0.011}$ & &0.028$^*_{\pm 0.002}$ & 0.693$^*_{\pm 0.024}$ & &0.030$^*_{\pm 0.002}$ & 0.631$^*_{\pm 0.019}$  \\
Transformer & 0.015$^*_{\pm 0.002}$ & 0.927$_{\pm 0.022}$ & &0.020$^*_{\pm 0.001}$ & 0.845$^*_{\pm 0.009}$ & &0.027$^*_{\pm 0.001}$ & 0.675$^*_{\pm 0.004}$ & &0.031$^*_{\pm 0.002}$ & 0.557$^*_{\pm 0.015}$  \\

Informer & 0.042$^*_{\pm 0.017}$ & 0.562$_{\pm 0.269}$ & &0.035$^*_{\pm 0.004}$ & 0.575$^*_{\pm 0.114}$ & &0.037$^*_{\pm 0.008}$ & 0.498$^*_{\pm 0.102}$ & &0.036$^*_{\pm 0.002}$ & 0.483$^*_{\pm 0.014}$  \\
iTransformer & 0.014$^*_{\pm 0.001}$ & 0.919$^*_{\pm 0.026}$ & &\underline{0.016}$_{\pm 0.003}$ & \underline{0.911}$_{\pm 0.025}$ & &\underline{0.018}$_{\pm 0.002}$ & \underline{0.853}$_{\pm 0.019}$ & &0.032$_{\pm 0.012}$ & 0.660$_{\pm 0.134}$  \\
Mamba & 0.019$_{\pm 0.008}$ & 0.870$_{\pm 0.116}$ & &0.021$_{\pm 0.007}$ & 0.845$^*_{\pm 0.047}$ & &0.023$^*_{\pm 0.003}$ & 0.799$^*_{\pm 0.035}$ & &0.026$_{\pm 0.006}$ & 0.682$_{\pm 0.079}$  \\
Performer & 0.013$_{\pm 0.001}$ & 0.942$_{\pm 0.014}$ & &0.021$^*_{\pm 0.004}$ & 0.849$^*_{\pm 0.041}$ & &0.019$_{\pm 0.003}$ & 0.823$^*_{\pm 0.011}$ & &0.030$_{\pm 0.008}$ & 0.669$_{\pm 0.079}$  \\
AttentionMixer & \textbf{0.011}$_{\pm 0.003}$ & \underline{0.962}$_{\pm 0.012}$ & &0.017$_{\pm 0.005}$ & 0.901$_{\pm 0.041}$ & &0.018$_{\pm 0.003}$ & 0.832$_{\pm 0.043}$ & &0.023$_{\pm 0.003}$ & \underline{0.766}$_{\pm 0.040}$  \\
FITS & 0.015$_{\pm 0.002}$ & 0.908$^*_{\pm 0.006}$ & &0.016$_{\pm 0.001}$ & 0.860$^*_{\pm 0.011}$ & &0.018$_{\pm 0.001}$ & 0.806$^*_{\pm 0.005}$ & &0.022$_{\pm 0.001}$ & 0.711$_{\pm 0.021}$  \\
FreTS & 0.016$_{\pm 0.007}$ & 0.920$_{\pm 0.049}$ & &0.017$_{\pm 0.003}$ & 0.887$^*_{\pm 0.013}$ & &0.021$_{\pm 0.004}$ & 0.810$^*_{\pm 0.032}$ & &\textbf{0.021}$_{\pm 0.002}$ & 0.757$_{\pm 0.022}$  \\
FBM & 0.015$_{\pm 0.003}$ & 0.883$^*_{\pm 0.045}$ & &0.018$^*_{\pm 0.002}$ & 0.854$^*_{\pm 0.017}$ & &0.020$_{\pm 0.001}$ & 0.781$^*_{\pm 0.015}$ & &0.023$_{\pm 0.003}$ & 0.702$_{\pm 0.028}$  \\
DeepFilter & \underline{0.012}$_{\pm 0.001}$ & \textbf{0.963}$_{\pm 0.006}$ & &\textbf{0.015}$_{\pm 0.001}$ & \textbf{0.927}$_{\pm 0.004}$ & &\textbf{0.018}$_{\pm 0.001}$ & \textbf{0.860}$_{\pm 0.018}$ & &\underline{0.022}$_{\pm 0.002}$ & \textbf{0.763}$_{\pm 0.053}$  \\
\hline
   \rowcolor{blue!8} \multicolumn{13}{l}{\textbf{Dataset: Jinan Station}} \\
AR & 0.106$^*_{\pm 0.000}$ & 0.744$^*_{\pm 0.000}$ & &0.118$^*_{\pm 0.000}$ & 0.668$^*_{\pm 0.000}$ & &0.123$^*_{\pm 0.000}$ & 0.620$^*_{\pm 0.000}$ & &0.130$^*_{\pm 0.000}$ & 0.571$^*_{\pm 0.000}$  \\
MA & 0.112$^*_{\pm 0.000}$ & 0.710$^*_{\pm 0.000}$ & &0.119$^*_{\pm 0.000}$ & 0.666$^*_{\pm 0.000}$ & &0.123$^*_{\pm 0.000}$ & 0.623$^*_{\pm 0.000}$ & &0.129$^*_{\pm 0.000}$ & 0.571$^*_{\pm 0.000}$  \\
ARIMA & 0.101$^*_{\pm 0.000}$ & 0.759$^*_{\pm 0.000}$ & &0.111$^*_{\pm 0.000}$ & 0.710$^*_{\pm 0.000}$ & &0.118$^*_{\pm 0.000}$ & 0.646$^*_{\pm 0.000}$ & &0.127$^*_{\pm 0.000}$ & 0.607$^*_{\pm 0.000}$  \\
LASSO & 0.029$^*_{\pm 0.000}$ & 0.742$^*_{\pm 0.000}$ & &0.030$^*_{\pm 0.000}$ & 0.717$^*_{\pm 0.000}$ & &0.031$^*_{\pm 0.000}$ & 0.680$^*_{\pm 0.000}$ & &0.033$^*_{\pm 0.000}$ & 0.636$^*_{\pm 0.000}$  \\
SVR & 0.061$^*_{\pm 0.000}$ & 0.640$^*_{\pm 0.000}$ & &0.065$^*_{\pm 0.000}$ & 0.599$^*_{\pm 0.000}$ & &0.056$^*_{\pm 0.000}$ & 0.679$^*_{\pm 0.000}$ & &0.050$^*_{\pm 0.000}$ & 0.706$^*_{\pm 0.000}$  \\
XGBoost & 0.031$^*_{\pm 0.000}$ & 0.903$^*_{\pm 0.000}$ & &0.033$^*_{\pm 0.000}$ & 0.883$^*_{\pm 0.000}$ & &0.036$_{\pm 0.000}$ & 0.853$^*_{\pm 0.000}$ & &0.038$^*_{\pm 0.000}$ & 0.816$^*_{\pm 0.000}$  \\
LSTM & 0.015$^*_{\pm 0.000}$ & 0.948$^*_{\pm 0.001}$ & &0.017$^*_{\pm 0.000}$ & 0.921$^*_{\pm 0.005}$ & &0.019$^*_{\pm 0.000}$ & 0.888$^*_{\pm 0.004}$ & &0.021$^*_{\pm 0.000}$ & 0.853$^*_{\pm 0.004}$  \\
GRU & 0.016$^*_{\pm 0.001}$ & 0.939$^*_{\pm 0.005}$ & &0.018$^*_{\pm 0.000}$ & 0.914$^*_{\pm 0.006}$ & &0.020$^*_{\pm 0.000}$ & 0.882$^*_{\pm 0.005}$ & &0.022$_{\pm 0.001}$ & 0.842$^*_{\pm 0.012}$  \\
Transformer & 0.016$_{\pm 0.004}$ & 0.955$_{\pm 0.023}$ & &0.017$^*_{\pm 0.002}$ & 0.929$^*_{\pm 0.014}$ & &0.024$^*_{\pm 0.005}$ & 0.837$^*_{\pm 0.057}$ & &0.025$^*_{\pm 0.002}$ & 0.813$^*_{\pm 0.013}$  \\
Informer & 0.018$^*_{\pm 0.002}$ & 0.917$^*_{\pm 0.013}$ & &0.025$_{\pm 0.008}$ & 0.851$^*_{\pm 0.041}$ & &0.022$^*_{\pm 0.000}$ & 0.813$^*_{\pm 0.003}$ & &0.026$^*_{\pm 0.002}$ & 0.761$^*_{\pm 0.021}$  \\
iTransformer & 0.012$_{\pm 0.000}$ & 0.971$^*_{\pm 0.005}$ & &0.014$_{\pm 0.002}$ & 0.948$_{\pm 0.028}$ & &0.018$_{\pm 0.002}$ & 0.906$_{\pm 0.023}$ & &\underline{0.020}$_{\pm 0.004}$ & 0.881$_{\pm 0.033}$  \\
Mamba & 0.015$_{\pm 0.004}$ & 0.957$_{\pm 0.023}$ & &0.017$^*_{\pm 0.006}$ & 0.905$^*_{\pm 0.085}$ & &0.018$^*_{\pm 0.006}$ & 0.922$^*_{\pm 0.015}$ & &0.020$_{\pm 0.002}$ & 0.883$_{\pm 0.017}$  \\
Performer & 0.013$_{\pm 0.004}$ & 0.948$_{\pm 0.037}$ & &0.014$_{\pm 0.001}$ & 0.950$_{\pm 0.004}$ & &0.019$_{\pm 0.003}$ & 0.895$_{\pm 0.018}$ & &0.023$_{\pm 0.006}$ & 0.871$_{\pm 0.038}$  \\
AttentionMixer & \underline{0.011}$_{\pm 0.004}$ & \underline{0.981}$_{\pm 0.012}$ & &\underline{0.012}$_{\pm 0.002}$ & \underline{0.967}$_{\pm 0.007}$ & &\underline{0.016}$_{\pm 0.002}$ & 0.912$_{\pm 0.035}$ & &0.027$_{\pm 0.010}$ & 0.840$_{\pm 0.039}$  \\
FITS & 0.013$_{\pm 0.003}$ & 0.965$^*_{\pm 0.006}$ & &0.015$_{\pm 0.002}$ & 0.941$^*_{\pm 0.002}$ & &0.017$_{\pm 0.002}$ & 0.917$^*_{\pm 0.007}$ & &0.019$_{\pm 0.002}$ & \underline{0.886}$_{\pm 0.005}$  \\
FreTS & 0.011$_{\pm 0.001}$ & 0.968$_{\pm 0.012}$ & &0.020$_{\pm 0.007}$ & 0.933$^*_{\pm 0.014}$ & &0.017$^*_{\pm 0.001}$ & \underline{0.921}$^*_{\pm 0.009}$ & &0.019$_{\pm 0.002}$ & 0.872$_{\pm 0.019}$  \\
FBM & 0.015$^*_{\pm 0.003}$ & 0.963$^*_{\pm 0.009}$ & &0.016$_{\pm 0.006}$ & 0.941$^*_{\pm 0.013}$ & &0.016$_{\pm 0.002}$ & 0.919$^*_{\pm 0.011}$ & &0.019$_{\pm 0.001}$ & 0.862$_{\pm 0.010}$  \\
DeepFilter & \textbf{0.010}$_{\pm 0.002}$ & \textbf{0.986}$_{\pm 0.003}$ & &\textbf{0.012}$_{\pm 0.000}$ & \textbf{0.970}$_{\pm 0.004}$ & &\textbf{0.015}$_{\pm 0.001}$ & \textbf{0.940}$_{\pm 0.007}$ & &\textbf{0.019}$_{\pm 0.003}$ & \textbf{0.890}$_{\pm 0.023}$  \\
    \bottomrule
    \end{tabular}
    \begin{tablenotes}
        \item \textit{Note}: The results are reported in $\mathrm{mean}_{\pm \mathrm{std}}$. The best and second best metrics are bolded and underlined, respectively. "*" marks the baselines that are significantly inferior to DeepFilter, with $p$-value $<0.05$ in the two-sample t-test.
    \end{tablenotes}
    \end{threeparttable}
    \end{table*} 

\subsubsection{Datasets} 
We use two industrial datasets from the ANRMN project's monitoring logs. The Hegang dataset contains 38,686 observations, and the Jinan dataset contains 38,687 observations, both with a 5-minute interval between observations. Each dataset includes 1,034 input variables and a quality variable (dose rate), as shown in~\autoref{dataTab}. The datasets are divided into training (70\%), validation (15\%), and testing (15\%) subsets and normalized using min-max scaling.

\subsubsection{Baselines} 
We include three categories of baselines: (1) Identification methods: AR, MA and ARIMA~\cite{Chandrakar2017}; (2) Statistical methods: Lasso Regression (LASSO)~\cite{Chan2017}, Support Vector Regression (SVR)~\cite{Tang2012}, Random Forest (RF)~\cite{rf}, and eXtreme Gradient Boosting (XGB)~\cite{xgb};
    (3) Deep methods: Long Short-Term Memory (LSTM)~\cite{Choi2020}, Gated Recurrent Unit (GRU)~\cite{zhang2018}, Transformer~\cite{Vaswani2017}, Informer~\cite{zhou2021informer}, Mamba~\cite{gu2024mamba}, Performer~\cite{Performer}, AttentionMixer~\cite{wang2024taiattentionmixer}, iTransformer~\cite{liuitransformer}. We also incorporate representative deep methods incorporating Fourier analysis, viz. FreTS~\cite{frets}, FITS~\cite{xu2023fits} and FBM~\cite{fbm}. 
    
\subsubsection{Implementation details} 

Experiments are conducted using the Adam optimizer~\cite{Kingma2015}. The hardware includes 2 Intel Xeon Platinum 8383C CPUs and 8 NVIDIA GeForce 1080Ti GPUs. Aligning with AttentionMixer~\cite{wang2024taiattentionmixer}, we employ a GRU decoder for transformer-based baselines to produce quality variable predictions. The learning rate is tuned within $\{0.001, 0.005, 0.01\}$, the batch size within $\{32, 64\}$, with 2 GF blocks and a historical window length of 16. Training is capped at 200 epochs with early stopping after 15 epochs without validation improvement. Test set performance metrics are then calculated and reported.
Monitoring accuracy is evaluated using the coefficient of determination (\(\mathrm{R}^2\)), which measures the proportion of variance in $y^\RH$ predictable from $\hat{y}^\RH$. We also incorporate the root mean squared error (RMSE) and the mean absolute error (MAE) as supplementary metrics to quantify prediction error magnitudes.

\subsection{Overall Performance}\label{sec:overall}

In this section, we compare the monitoring accuracy of DeepFilter with baseline models, as shown in \autoref{comparativeTab}. Key observations are as follows: 
\ding{182}~\textbf{Identification models demonstrate limited accuracy in process monitoring.} Primarily designed to capture linear autocorrelations, they exhibit limitations in modeling the nonlinear patterns that are prevalent in monitoring logs. Statistical models, which integrate external factors like meteorological conditions, perform competitively in short-term monitoring. Nonlinear estimators, particularly XGBoost, outperform linear models and achieve results comparable to traditional deep learning models like LSTM and GRU.
\ding{183}~\textbf{Deep models achieve the best performance among baselines.} Specifically, Transformer-based methods display varying accuracy contingent upon their temporal fusion mechanisms. The canonical Transformer using self-attention shows suboptimal performance, while modifications to the temporal fusion mechanisms, such as iTransformer and AttentionMixer, significantly enhance performance. This underscores the limitations of self-attention in modeling monitoring logs and the need to refine the temporal fusion mechanisms to adapt the Transformer for process monitoring.
\ding{184}~\textbf{DeepFilter demonstrates the best overall performance across different datasets.} Its superior accuracy is due to the efficient filtering layer, which effectively captures long-term discriminative patterns, aiding in the understanding of monitoring logs.

\begin{figure*}
    \subfigure[Performance comparison on the Hegang dataset.]{\includegraphics[width=\linewidth]{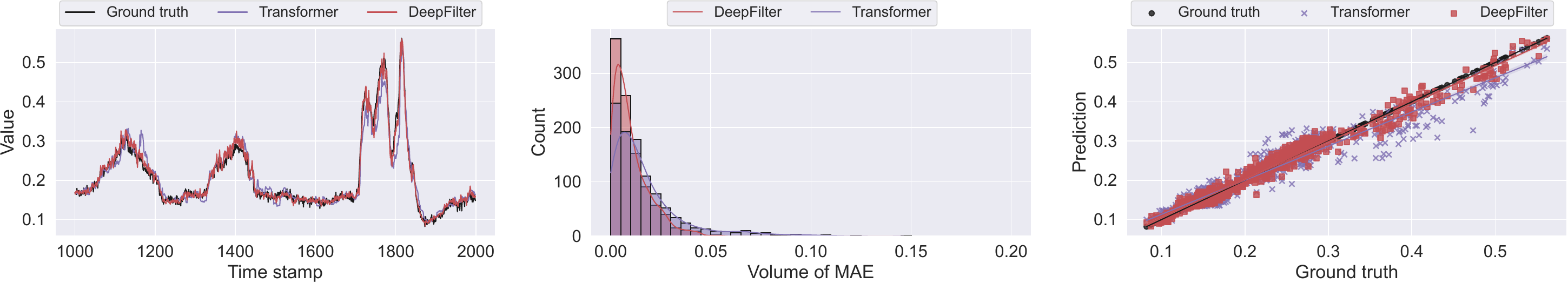}}
    \subfigure[Performance comparison on the Jinan dataset.]{\includegraphics[width=\linewidth]{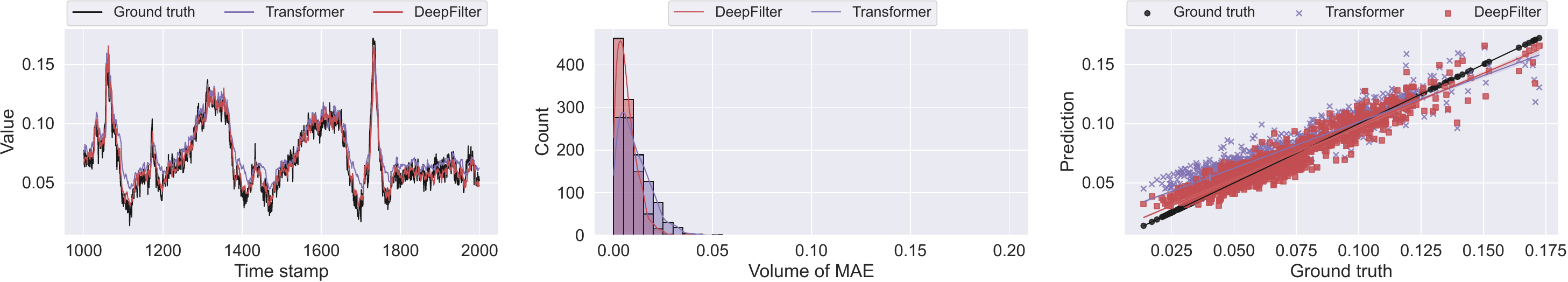}}
    \caption{In-depth comparison on the monitoring performance of DeepFilter and Transformer.}\label{fig:compare}
\end{figure*}

\begin{table*}
\caption{Ablation study results.}\label{tab:ablation}
\centering\renewcommand{\arraystretch}{1}\small\setlength{\tabcolsep}{4.5mm}
\begin{tabular}{l|cc|cc|cc}
\hline
      & \multicolumn{6}{c}{Hegang Station} \\
\hline
 Model & RMSE & $\Delta$RMSE & MAE & $\Delta$MAE & R$^2$ & $\Delta$R$^2$   \\ \hline 
 DeepFilter & \textbf{0.016}$_{\pm 0.001}$ & - & \textbf{0.012}$_{\pm 0.001}$ & - & \textbf{0.963}$_{\pm 0.006}$ & - \\
 w/o efficient filtering layer & 0.022$_{\pm 0.003}$ & \textcolor{abl}{37.50\%$\uparrow$} & 0.015$_{\pm 0.002}$ & \textcolor{abl}{25.00\%$\uparrow$} & 0.927$_{\pm 0.019}$ & \textcolor{abl}{3.73\%$\downarrow$} \\
 w/o fully connected layer & 0.019$_{\pm 0.002}$ & \textcolor{abl}{18.75\%$\uparrow$} & 0.013$_{\pm 0.001}$ & \textcolor{abl}{8.33\%$\uparrow$} & 0.946$_{\pm 0.013}$ & \textcolor{abl}{1.76\%$\downarrow$} \\ 
 w/o FFT & 0.017$_{\pm 0.001}$ & \textcolor{abl}{6.25\%$\uparrow$} & 0.012$_{\pm 0.000}$ & - & 0.955$_{\pm 0.003}$ & \textcolor{abl}{0.83\%$\downarrow$} \\
\hline
 & \multicolumn{6}{c}{Jinan Station} \\
\hline
Model & RMSE & $\Delta$RMSE & MAE & $\Delta$MAE & R$^2$ & $\Delta$R$^2$   \\ \hline 
 DeepFilter                    & \textbf{0.012}$_{\pm 0.001}$ & - & \textbf{0.010}$_{\pm 0.002}$ & - & \textbf{0.986}$_{\pm 0.003}$ & - \\
 w/o efficient filtering layer & 0.019$_{\pm 0.001}$ & \textcolor{abl}{58.33\%$\uparrow$} & 0.014$_{\pm 0.002}$ & \textcolor{abl}{40.00\%$\uparrow$} & 0.968$_{\pm 0.002}$ & \textcolor{abl}{1.82\%$\downarrow$} \\
 w/o fully connected layer                 & 0.020$_{\pm 0.001}$   & \textcolor{abl}{66.66\%$\uparrow$} & 0.013$_{\pm 0.001}$ & \textcolor{abl}{30.00\%$\uparrow$} & 0.966$_{\pm 0.005}$ & \textcolor{abl}{2.02\%$\downarrow$} \\ 
 w/o FFT & 0.019$_{\pm 0.004}$ & \textcolor{abl}{58.33\%$\uparrow$} & 0.012$_{\pm 0.002}$ & \textcolor{abl}{20.00\%$\uparrow$} & 0.968$_{\pm 0.012}$ & \textcolor{abl}{1.82\%$\downarrow$}  \\
 \hline
\end{tabular}
\end{table*}

To qualitatively evaluate model performance, we compare DeepFilter with the canonical Transformer across three key aspects, as shown in \autoref{fig:compare}. The main observations are as follows. \ding{182}~\textbf{DeepFilter provides more accurate and stable predictions.} The left panels compare the predicted series against the ground truth. DeepFilter closely follows the true signal across both steady and dynamic regions, particularly during abrupt fluctuations. In contrast, the Transformer often fails to predict sharp variations, such as the spikes around timestamp 1750. \ding{183}~\textbf{DeepFilter yields a more compact and consistent error distribution.} The middle panels present the MAE distributions. DeepFilter produces a concentrated error distribution with lower variance, demonstrating improved reliability for real-time monitoring tasks. \ding{184}~\textbf{DeepFilter achieves a higher goodness of fit to the ground truth.} The right panels show scatter plots of predicted versus actual values. DeepFilter's predictions form a tight diagonal cluster, signifying strong alignment with the ground truth. Conversely, the Transformer exhibits greater dispersion and several outliers.

\subsection{Ablation Analysis}\label{sec:ab}
In this section, we examine the contributions of the efficient filtering layer and the fully connected layer to the overall performance of DeepFilter. The results in \autoref{tab:ablation} reveal that both components are essential, as removing either leads to notable performance declines. For example, ablating the efficient filtering layer increases the MAE by 25\% on the Hegang dataset, while removing the fully connected layer reduces the R$^2$ by 2.02\% on the Jinan dataset. Additionally, we replace the FFT for representation transformation with a convolution using a randomly initialized kernel of the same length as the input sequence. This change results in a slight but consistent performance drop, highlighting the efficacy of FFT in capturing long-term discriminative patterns as orthogonal components, as elaborated in Section~\ref{sec:theory}.


\subsection{Efficiency Analysis}\label{sec:complex}
\begin{figure*}
\centering
\subfigure[Efficiency metrics given varying historical lengths on the NVIDIA GeForce 1080Ti GPU.]{
\includegraphics[width=0.98\linewidth]{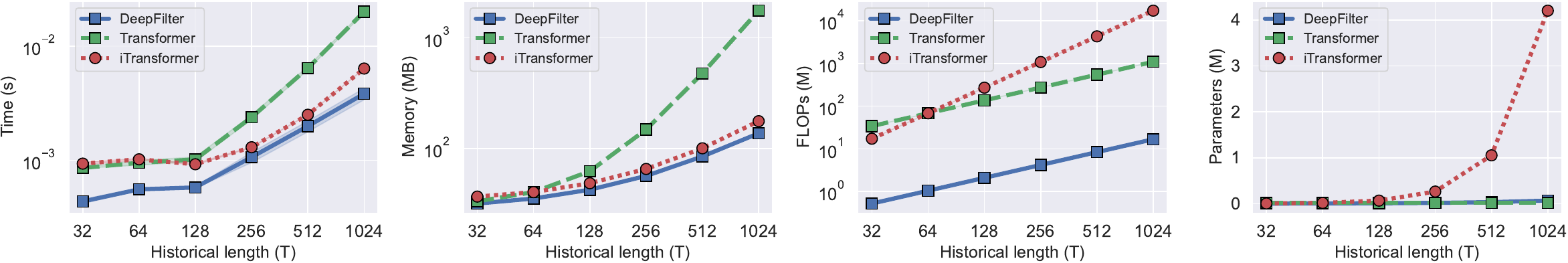}}
\subfigure[Efficiency metrics given varying hidden dimensions on the NVIDIA GeForce 1080Ti GPU.]{
\includegraphics[width=0.98\linewidth]{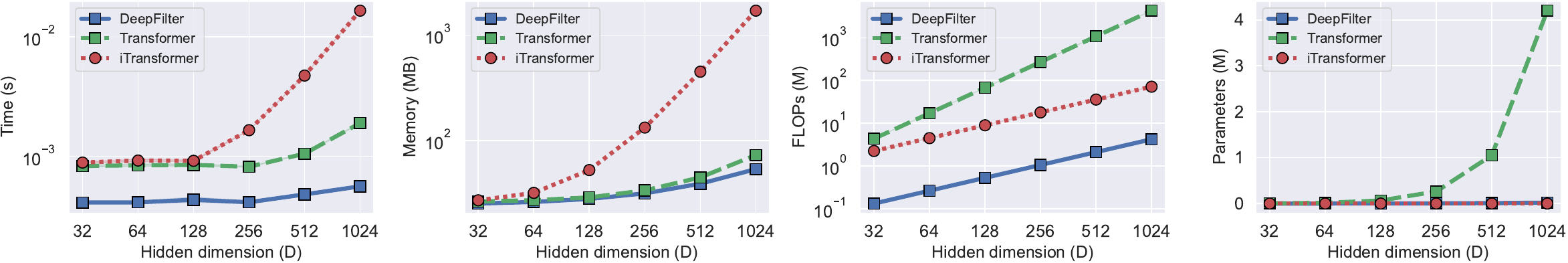}}
\caption{Efficiency analysis results given varying settings, with solid lines for mean values of 10 trials and shaded areas for 90\% confidence intervals. The default values of T, D and batch size are 16, 32, and 64, respectively.}\label{fig:speed}
\end{figure*}

In this section, we empirically validate the efficiency of DeepFilter by benchmarking it against the Transformer~\cite{Vaswani2017} and iTransformer~\cite{liuitransformer} baselines. The evaluation focuses on the temporal fusion layer where they differ (e.g., the self-attention layer in Transformer and the efficient filtering layer in DeepFilter), by tasking each to transform a sequence \((\mathbf{x}_1,...,\mathbf{x}_\T)\) into another sequence \((\mathbf{z}_1,...,\mathbf{z}_\T)\) of equal length \(\T\) and feature dimension \(\D\). The results are presented in \autoref{fig:speed} with key observations as follows. 
\ding{182}~\textbf{Transformer-based models demonstrate significant scaling in complexity as historical length and hidden dimension increase.} Specifically, for inference time and memory usage, Transformer shows the most rapid increases with growing historical length, while iTransformer is most sensitive to increases in hidden dimension. For parameter counts and FLOPs, Transformer's cost scales fastest with hidden dimension, whereas iTransformer scales fastest with historical length. These trends align with our analysis in \autoref{tab:complex} where they exhibit quadratic complexity scaling to different parameters.
\ding{183}~\textbf{DeepFilter achieves consistently lower complexity across all evaluated metrics.} For all complexity metrics in \autoref{fig:speed}, DeepFilter demonstrates minimal growth with respect to both historical length and hidden dimension. This is consistent with our analysis in \autoref{tab:complex}, where DeepFilter avoids the quadratic scaling issues in the Transformer-based models. 
\ding{184}~\textbf{DeepFilter's efficiency is suitable for real-time process monitoring.} For example, with a historical length of 1024, DeepFilter achieves inference times below 0.01 seconds, FLOPs under 50M, and memory usage below 200 MB, confirming its practical applicability under stringent latency and resource constraints.


\section{Conclusion}\label{sec:conclusion}

In this paper, we introduced DeepFilter, an adaptation of the Transformer architecture specifically for process monitoring. By replacing the self-attention layer with an efficient filtering layer, DeepFilter excels at capturing long-term discriminative patterns inherent in monitoring logs while significantly reducing computational complexity. Experiments on process monitoring datasets demonstrate that DeepFilter outperforms diverse models in both accuracy and efficiency, effectively meeting the stringent demands of modern process monitoring.

\noindent\textbf{Limitations and Future Work.} The current implementation of DeepFilter utilizes the FFT for obtaining frequency-domain representation.
While FFT is efficient and effective for stationary or quasi-stationary monitoring data, it inherently assumes temporal stationarity and uniform frequency composition across different time intervals. This assumption limits the model's ability to capture transient or non-stationary dynamics with evolving frequency composition in industrial monitoring logs. Consequently, DeepFilter may exhibit degraded performance when encountering monitoring logs characterized by rapid shifts or non-stationary signal components.
Future work could address this limitation by exploring more flexible transformation approaches—such as wavelet transforms, short-time Fourier transforms, or adaptive time–frequency decomposition techniques—that can better localize temporal variations and handle non-stationary process behaviors.

\bibliographystyle{ieeetr}
\footnotesize{\bibliography{Bibliography/abbr,Bibliography/main,causal,Bibliography/main2,Bibliography/main3}}
\clearpage
\normalsize
\appendix

\subsection{Parameter Sensitivity Study}\label{sec:param}

In this section, we investigate the impact of key hyperparameters on DeepFilter's performance, including the number of global filtering blocks (K), historical length (T), hidden dimension (D), and batch size (B).

\begin{figure}[h]
    \centering
    \subfigure[Performance with varying numbers of blocks (K).]{\includegraphics[width=0.98\linewidth]{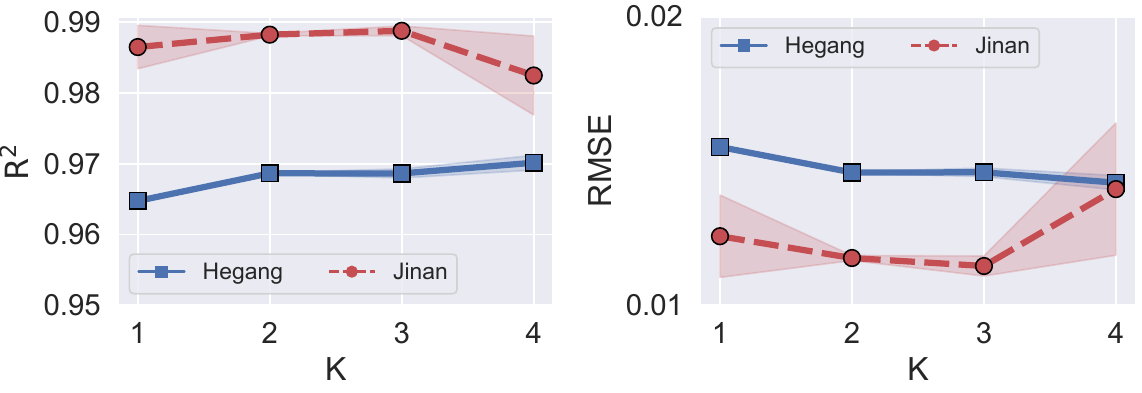}}
    \subfigure[Performance with varying settings of window length (L).]{\includegraphics[width=0.98\linewidth]{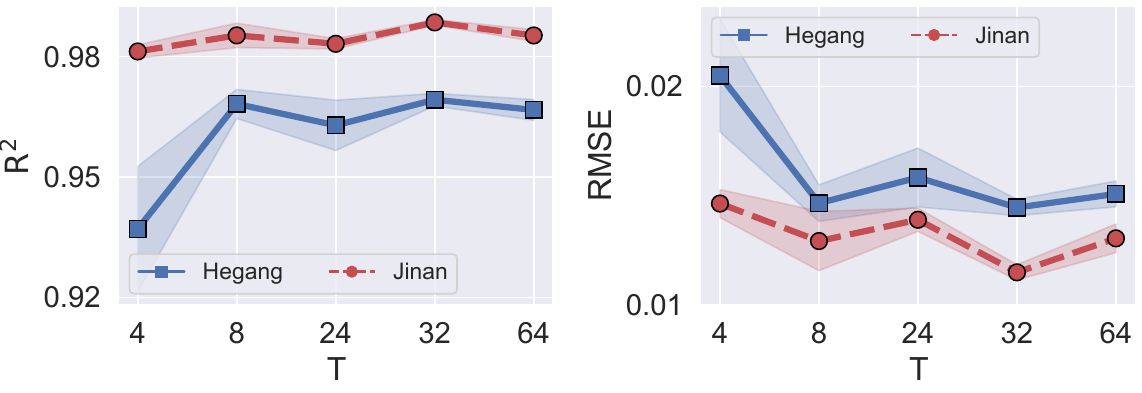}}
    \subfigure[Performance with varying hidden dimensions (D).]{\includegraphics[width=0.98\linewidth]{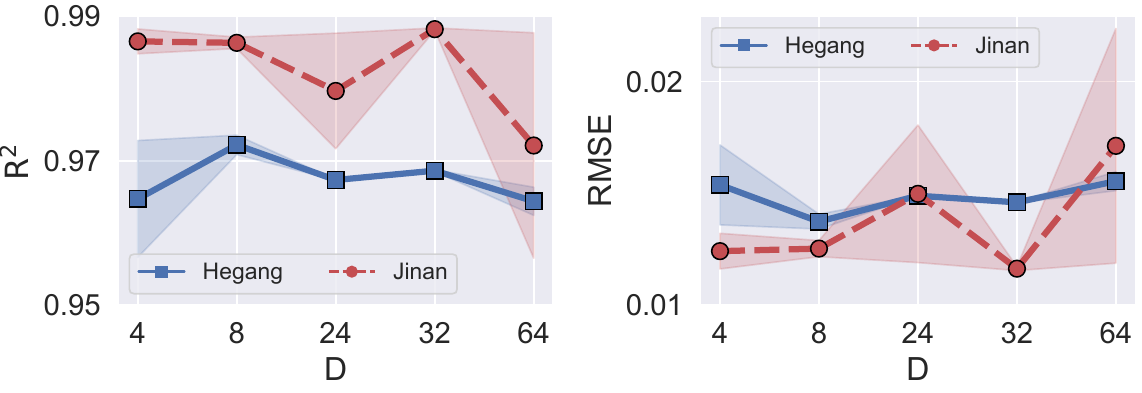}}
    \subfigure[Performance with varying settings of batch size.]{\includegraphics[width=0.98\linewidth]{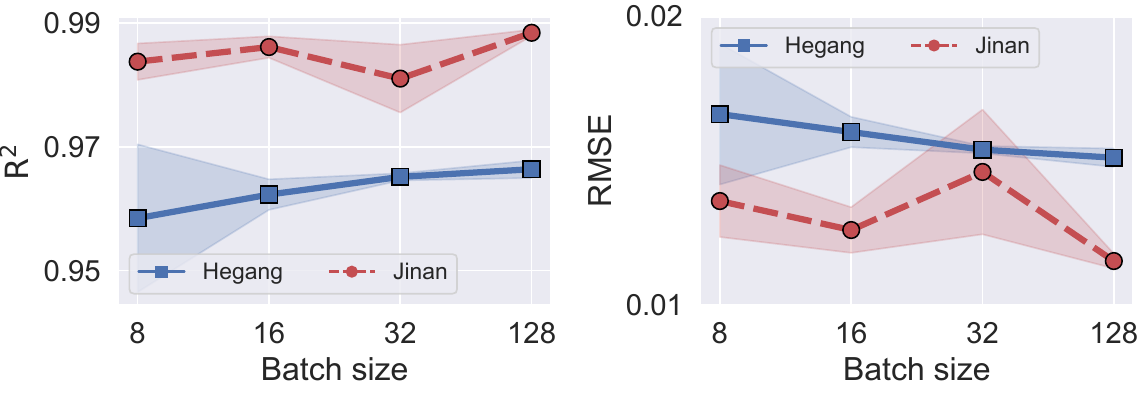}}
    \caption{Parameter sensitivity on Hegang and Jinan datasets, where the dark areas indicate the 90\% confidence intervals.}
    \label{fig:sensi}
\end{figure}

The results are summarized in \autoref{fig:sensi} with key observations as follows:
\ding{182}~\textbf{DeepFilter does not necessitate a deep stack of blocks.} As illustrated in \autoref{fig:sensi} (a), utilizing 1-2 blocks already yields promising results. While adding more blocks can incrementally improve performance, there is a risk of performance degradation, potentially due to overfitting and optimization challenges.
\ding{183}~\textbf{The performance improves with the increasing inclusion of historical monitoring data.} In \autoref{fig:sensi} (b), we observe that performance is suboptimal at T=4 but significantly enhances at T=8. Extending the window length further can lead to improved monitoring accuracy, indicating the value of incorporating sufficient historical context in the model.
\ding{184}~\textbf{The relationship between the number of hidden dimensions and model performance is not significant.} In \autoref{fig:sensi} (c), a smaller dimension appears sufficient to effectively model the monitoring logs at both stations, suggesting that the logs contain a high degree of redundancy. Finally, enlarging batch sizes generally improves performance in \autoref{fig:sensi} (d), which suggests the potential benefit of increasing the batch size to further enhance model performance.

\begin{figure}
    \centering
    \subfigure[Sensitivity with the numbers of blocks (K) and hidden dimension (D).]{\includegraphics[width=\linewidth]{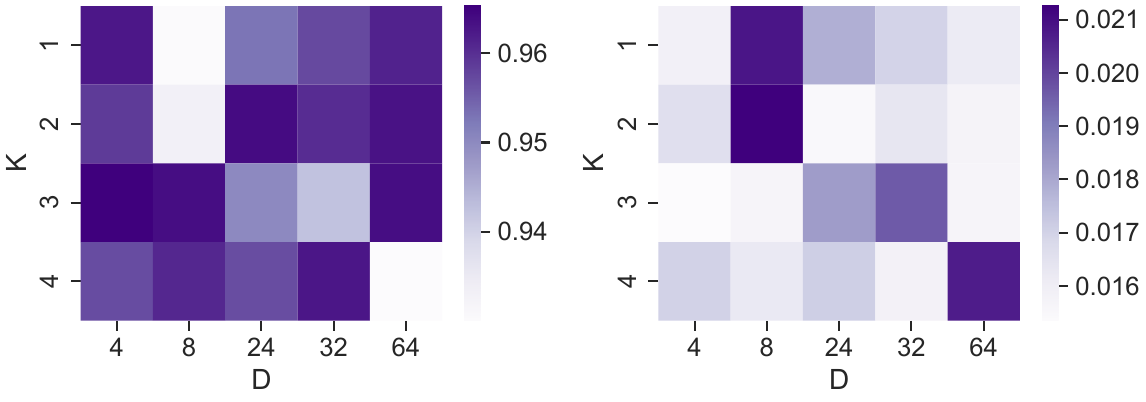}}
    \subfigure[Sensitivity with the historical length (T) and hidden dimension (D).]{\includegraphics[width=\linewidth]{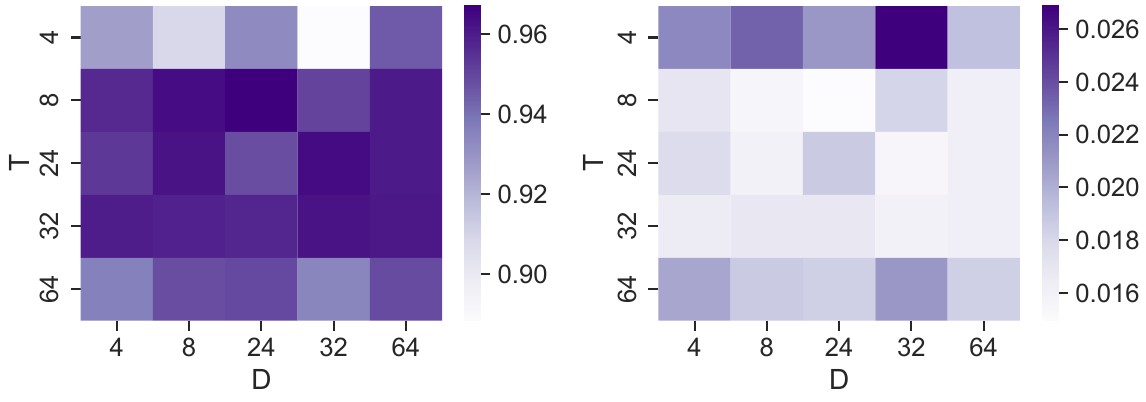}}
    \caption{Compositional parameter sensitivity on the Hegang dataset, where the metrics are R$^2$ (in left panels) and RMSE (in right panels).}
    \label{fig:sensi2}
\end{figure}

We further investigate the simultaneous variation of two hyperparameters and present the results in \autoref{fig:sensi2}. The key observations are summarized as follows.
\ding{182}~\textbf{For cases with a small number of blocks ($\mathrm{K}\leq3$), increasing the hidden dimension can potentially enhance performance.} A larger hidden dimension boosts the model's capacity to fit the monitoring data and improves fitting performance. However, for a large number of blocks ($\mathrm{K}=4$), excessively increasing the hidden dimension to 64 results in a significant performance decline due to overfitting, which poses a risk of poor generalization on the test set. This phenomenon is attributed to overfitting, where excessively large models exhibit poor generalization performance on the test set.
\ding{183}~\textbf{A similar trend is observed in the interaction between window length and hidden dimension.} When the window length is small ($\mathrm{T}=1$), the optimal hidden dimension with the highest $\mathrm{R}^2$ is 64. As the window length increases to 8, 24, and 32, the optimal $\mathrm{D}$ ranges from 24 to 32. When the window length reaches $\mathrm{T}=64$, the optimal $\mathrm{D}$ decreases to 24. This is because an overly small model cannot effectively fit the monitoring logs, while an overly large model risks overfitting. This creates a trade-off between window length and hidden dimension, affecting the model capacity.

\subsection{Dataset Difference Comparison}

\begin{figure}
    \centering
    \includegraphics[width=\columnwidth]{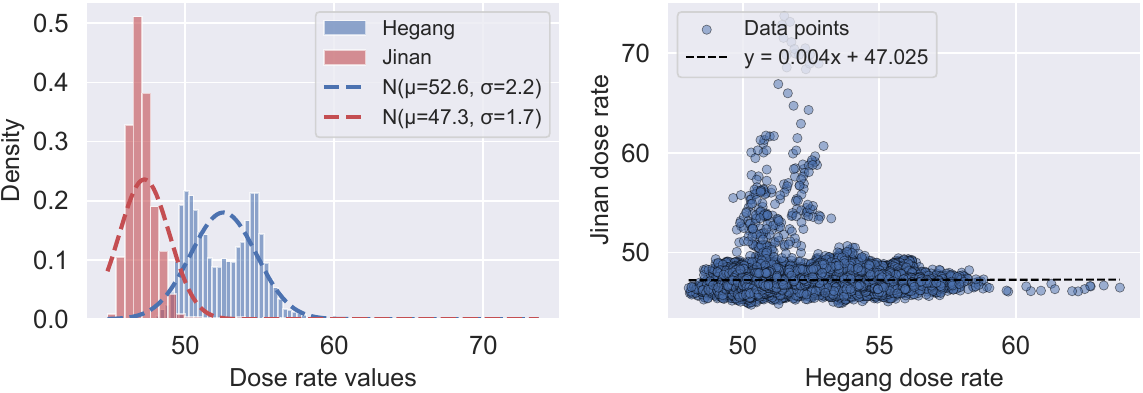}
    \caption{Comparison of dose rate distributions between the Hegang and Jinan datasets. The left panel illustrates Gaussian-fitted dose rate distributions for both stations, while the right panel presents their linear correlation.}
    \label{fig:diff}
\end{figure}

To rigorously assess the differences between datasets collected at two monitoring stations, we conduct a case study on the dose rate distribution, which serves as the key quality variable affecting monitoring performance. \autoref{fig:diff} visualizes the dose rate distributions for the Hegang and Jinan datasets, together with Gaussian fits and linear correlation analysis. The main findings are summarized as follows.
\ding{182}~\textbf{The dose rate distributions of the two datasets exhibit significant statistical differences.}
The Hegang dataset shows a mean of 52.6 and a standard deviation of 2.2, whereas the Jinan dataset records a mean of 47.3 and a standard deviation of 1.7. A two-sample t-test yields a p-value below 0.001, confirming that the two distributions differ significantly. Moreover, linear regression of the Jinan dose rate on the Hegang dose rate yields a slope of only 0.004 and an R-squared value near zero, indicating negligible linear correlation between the two datasets.
\ding{183}~\textbf{The Jinan dataset presents a more predictable and well-behaved label distribution.}
The Jinan dose rate follows an approximately Gaussian distribution with high density around the mean, implying a regular label distribution that is easier to model. In contrast, the Hegang distribution deviates notably from Gaussianity, with sparse observations near the mean and heavier tails, suggesting an irregular or complex label distribution that is harder to model. Moreover, the narrower variance of the Jinan dataset further enhances its predictability and modeling efficiency. These properties collectively render the dose rates in the Jinan dataset easier to model, consistent with observations from \autoref{comparativeTab} where the process monitors exhibit better performance on the Jinan dataset.

\end{document}